\pdfoutput=1
\PassOptionsToPackage{table}{xcolor}
\documentclass[11pt]{article}

\usepackage[final]{acl}

\usepackage{times}
\usepackage{latexsym}
\usepackage{amsmath}
\usepackage{subcaption}
\usepackage{wrapfig}
\usepackage{multirow}
\usepackage{booktabs}
\usepackage{xcolor}

\usepackage[T1]{fontenc}

\usepackage[utf8]{inputenc}

\usepackage{microtype}

\usepackage{inconsolata}

\usepackage{graphicx}
\usepackage{textcomp}
\newcommand{\method}{\textsc{Mlan}}
\newcommand{\methodbf}{\textbf{\textsc{Mlan}}}
\newcommand{\methodfull}{\text{\textbf{M}ultimodal \textbf{LAN}guage-based instruction tuning}}  
\NewDocumentCommand{\llama}{m o}{%
  \begingroup
  \def\modelname{}%
  \ifnum#1=3 \def\modelname{Llama-3.2-3B}\fi
  \ifnum#1=8 \def\modelname{Llama-3.1-8B}\fi
  \modelname
  \IfValueT{#2}{-Instruct}%
  \endgroup
}

\newcommand{\llavamix}{\text{MIX-LLaVA-1.5}}
\newcommand{\cambrianmix}{\text{MIX-Cambrian-1}}
\newcommand{\languageonly}{\text{text-only}}
\newcommand{\languageonlystart}{\text{Text-only}}
\newcommand{\languageonlycap}{\text{Text-Only}}
\newcommand{\visionlanguage}{\text{vision-language}}

\newcommand{\visionlanguagecap}{\text{Vision-Language}}
\newcommand{\totaltraininstances}{\text{186,000}}  
\newcommand{\supernatural}{\text{Super-NaturalInstructions}}  
\newcommand{\visionflan}{\text{Vision-Flan}}  
\newcommand{\numevaltasks}{\text{12}}  
\newcommand{\numtextevaltasks}{\text{7}}  
\newcommand{\numvisionevaltasks}{\text{5}}  

\newif\ifshowcomments
\showcommentsfalse 

\usepackage{color}

\newcommand{\nick}[1]{\ifshowcomments\textcolor{red}{{nick: #1}}\fi}

\title{\method: Language-Based Instruction Tuning Preserves and Transfers Knowledge in Multimodal Language Models}

\author{
  {\bf Jianhong Tu}\textsuperscript{1}\thanks{\ Equal contribution}, 
  {\bf Zhuohao Ni}\textsuperscript{2}\footnotemark[1], 
  {\bf Nicholas Crispino}\textsuperscript{1}, 
  {\bf Zihao Yu}\textsuperscript{1}, 
  {\bf Michael Bendersky}\textsuperscript{3},\\
  {\bf Beliz Gunel}\textsuperscript{3}, 
  {\bf Ruoxi Jia}\textsuperscript{4}, 
  {\bf Xin Liu}\textsuperscript{5}, 
  {\bf Lingjuan Lyu}\textsuperscript{6}, 
  {\bf Dawn Song}\textsuperscript{7}, 
  {\bf Chenguang Wang}\textsuperscript{1}\thanks{Corresponding author} \\
  \textsuperscript{1}Washington University in St. Louis \quad
  \textsuperscript{2}The University of British Columbia \quad
  \textsuperscript{3}Google Research \\
  \textsuperscript{4}Virginia Tech \quad
  \textsuperscript{5}University of California, Davis \quad
  \textsuperscript{6}Sony AI \quad
  \textsuperscript{7}University of California, Berkeley \\
  \texttt{\{jianhong.t, ncrispino, yu.zihao, chenguangwang\}@wustl.edu} \quad
  \texttt{peterni@student.ubc.ca} \\
  \texttt{\{bemike, bgunel\}@google.com} \quad
  \texttt{ruoxijia@vt.edu} \quad
  \texttt{xinliu@ucdavis.edu} \\
  \texttt{Lingjuan.Lv@sony.com} \quad
  \texttt{dawnsong@berkeley.edu}
}

\begin{document}
\maketitle

\begin{abstract}
We present a novel visual instruction tuning strategy to improve the zero-shot task generalization of multimodal large language models by building a firm \languageonly\ knowledge base.
Existing work lacks sufficient experimentation on the importance of each modality in the instruction tuning stage, often using a majority of \visionlanguage\ data while keeping \languageonly\ data limited and fixing mixtures of modalities.
By incorporating diverse \languageonly\ data in the visual instruction tuning stage, we vary \visionlanguage\ data in various controlled experiments to investigate the importance of modality in visual instruction tuning.
Our comprehensive evaluation shows that the text-heavy instruction tuning approach is able to perform on-par with traditional vision-heavy mixtures on both modalities across \numevaltasks\ general datasets while using as low as half the total training tokens.
We find that simply increasing sufficiently diverse \languageonly\ data enables transfer of instruction following ability and domain knowledge across modalities while being more efficient than the \visionlanguage\ approach.

\end{abstract}

\begin{figure*}[htbp]
    \centering
    \begin{subfigure}[b]{0.425\textwidth}
        \centering
        \includegraphics[width=\linewidth]{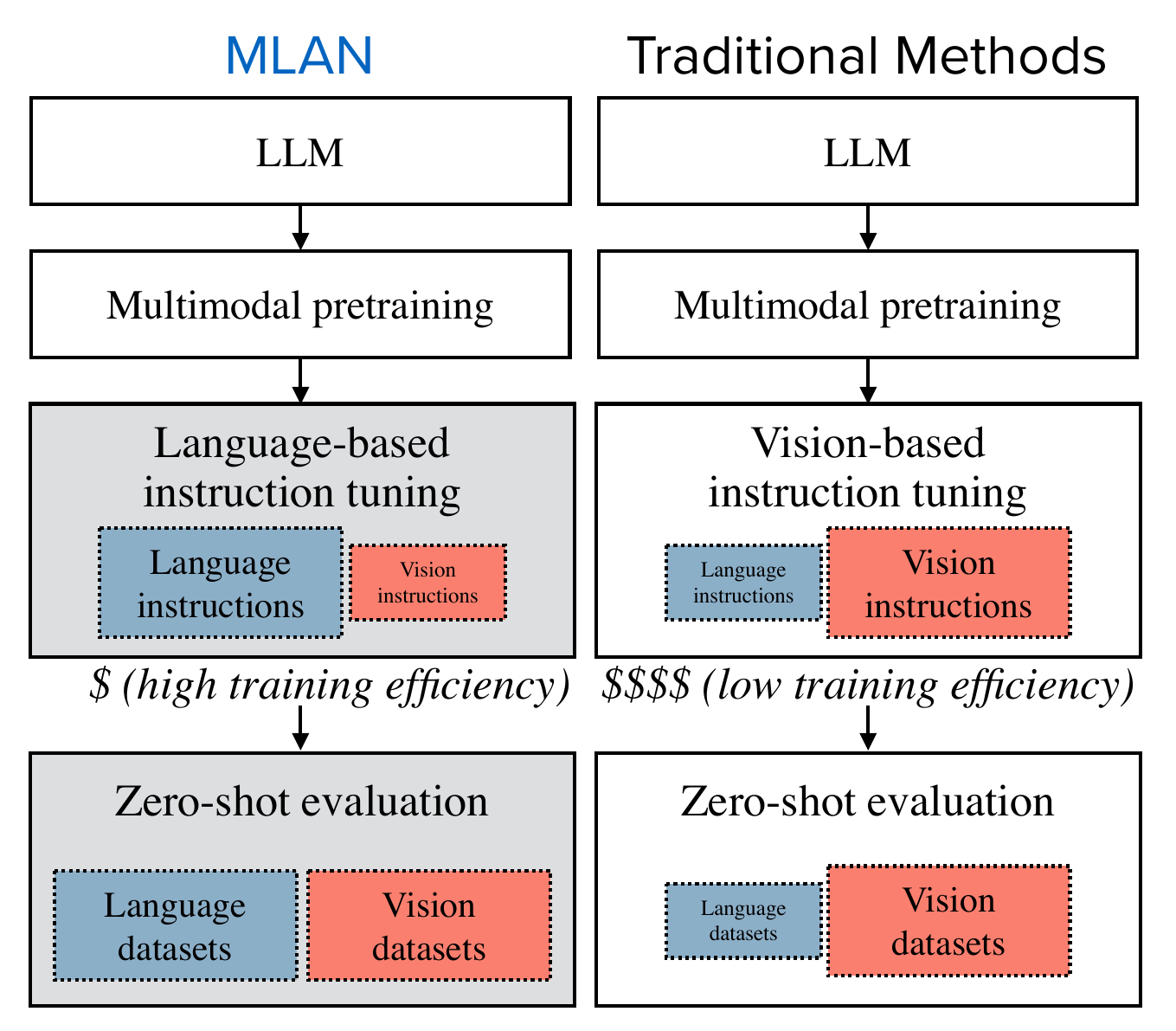}
        \vspace{0.015mm}
        \caption{ 
        Comparison of \method\ with standard visual instruction tuning.
        }
        \label{fig:approach}
    \end{subfigure}
    \hfill
    \begin{subfigure}[b]{0.535\textwidth}
        \centering
        \includegraphics[width=\linewidth]{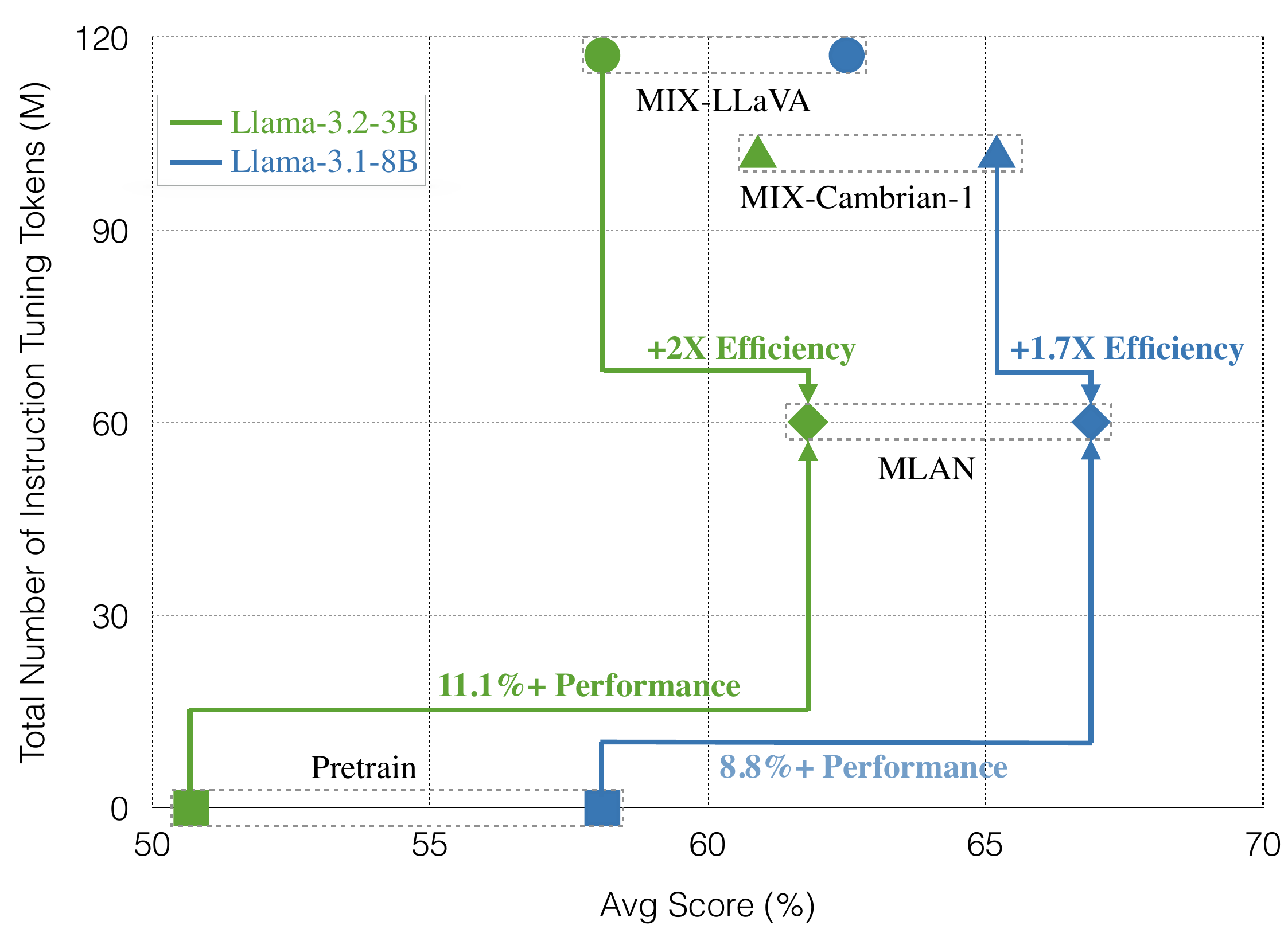}
        \caption{
        Main results on evaluation tasks, averaged over \languageonly\ and \visionlanguage\ performance.
        }
        \label{fig:results}
    \end{subfigure}
    
    \caption{
    Overview of \method. 
    (a) \method\ represents a shift in perspective towards text during instruction tuning. After vision-language pretraining, we include diverse \languageonly\ data in our instruction tuning mixture spanning many tasks. We emphasize including \languageonly\ data to show the transferability of instruction tuning across modalities.
    For evaluation, we select ample \languageonly\ and \visionlanguage\ datasets, allowing us to compare performance changes across modalities.
    (b) We evaluate \method\ on two pretrained multimodal models based on \llama{3} and \llama{8} across unseen language and vision benchmarks, achieving comparable performance at higher training efficiency (up to almost 2x as efficient compared to standard vision-heavy instruction tuning) with our language-based approach.
    }
    \label{fig:overview}
\end{figure*}

\section{Introduction}
Multimodal large language models (MLLMs) have advanced and enabled a wide range of \visionlanguage\ tasks such as visual question answering and image captioning~\citep{llava, flamingo, blip2, vila, baiQwen25VLTechnicalReport2025}. 
Their zero-shot generalization ability to unseen tasks has the potential to further revolutionize broader real-world applications~\citep{palme, zhu2023minigpt, llava-med}.
To construct MLLMs, vision-language pretraining is performed on a large scale with image-text data, aligning the modalities before visual instruction tuning aligns the model with human preferences~\citep{llava1.5, nvlm, vila}.
The importance of strong vision-language pretraining is established, with more data resulting in greater improvements in instruction-following abilities and downstream performance~\citep{mm1, zhangMM15MethodsAnalysis2024}.
However, current visual instruction tuning practices overwhelmingly rely on image-text pairs and large-scale vision-language datasets. This emphasis introduces a significant distributional shift from the language-rich corpora used during pretraining, often degrading the model’s general language understanding and leading to catastrophic forgetting of core knowledge~\citep {zhang2024wings}.
Given the similarity in instruction tuning data across modalities and the strong modality alignment achieved with vision-language pretraining, we believe \languageonly\ data is underutilized in existing training mixtures.  
Additionally, various design choices regarding the instruction tuning dataset composition with respect to modalities are underexplored.

In this work, we introduce \methodbf\ (\methodfull), a new perspective in vision instruction tuning that treats language as the primary way to unlock knowledge during instruction tuning (Figure~\ref{fig:overview}). Our key insight is that instruction-following abilities and domain knowledge, once acquired through diverse language-only tasks, can generalize across modalities with minimal vision-language supervision.
By grounding vision capabilities in a small number of targeted image-text examples, we maintain high performance across both vision and text tasks while significantly reducing training costs.
Specifically, with \method\ we unlock vision instruction following abilities by teaching a pretrained model to execute \languageonly\ instructions and then complementing the dataset with a relatively small portion of \visionlanguage\ examples in a domain adaptation fashion. 
To demonstrate \method's effectiveness, we pretrain MLLMs over a variety of settings based on \llama{3}~\citep{meta2024llama32} and \llama{8}~\citep{grattafiori2024llama3}, following the state of the art multimodal training mechanism~\citep{llava, llava1.5}, varying only the dataset. 
We then apply \method\ to the MLLMs and observe the following key insights over both models on average during evaluation on \numevaltasks\ comprehensive benchmarks across language and vision modalities. 
(1) Compared with the traditional vision-heavy finetuning approaches of LLaVA~\citep{llava1.5} and Cambrian-1~\citep{cambrian-1}, our models finetuned with \method\ demonstrate a matching or better performance on downstream \visionlanguage\ tasks while seeing less than half of the images and consistently showing better \languageonly\ performance. We show that text-only data is imperative to obtain world knowledge and understanding of complex instructions, even in the vision domain.
(2) \languageonlystart\ instruction tuning is more cost-effective. The rich and dense information compensates for the limited diversity in public vision datasets, allowing for superior performance while reducing the total number of processed training tokens by half.
(3) Neither language nor vision alone is enough for a generalist MLLM. Our experiments show that while instruction following abilities may transfer across modalities, their impact on the other modality is limited: certain \visionlanguage\ tasks do not benefit from \languageonly\ tuning and \visionlanguage\ tuning can result in severe degradation of language abilities. However, mixing bi-modal data, even at a small percentage, leads to surprising performance boosts and achieves the best results in both modalities. We hope our findings will foster future research on language-centered training and instruction tuning, paving the way for fundamental advancements in large MLLMs.

\label{training-task-selection}

\begin{figure*}[htbp]
    \centering
    \begin{subfigure}[b]{0.55\textwidth}
        \centering
        \includegraphics[width=\linewidth]{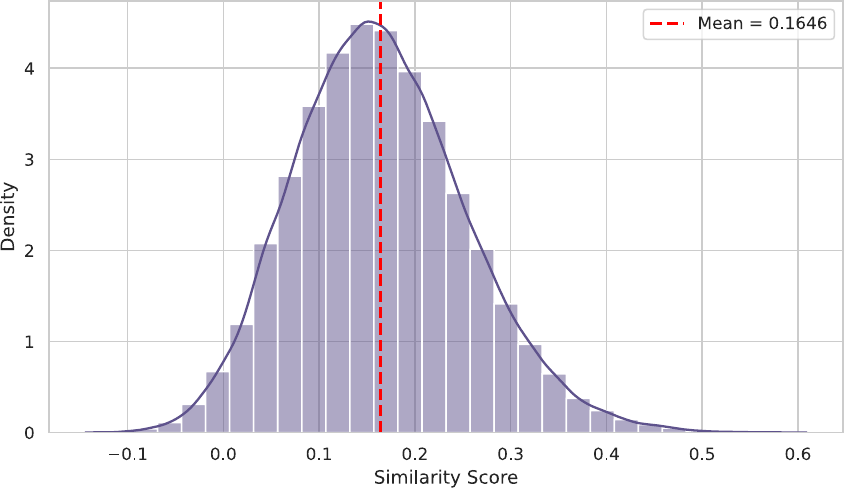}
        \caption{ 
            Distribution of cross-modal similarity scores between modalities with a non-negative mean by z-test ($p<0.001$).
        }
        \label{fig:approach}
    \end{subfigure}
    \hfill
    \begin{subfigure}[b]{0.4\textwidth}
        \centering
        \includegraphics[width=\linewidth]{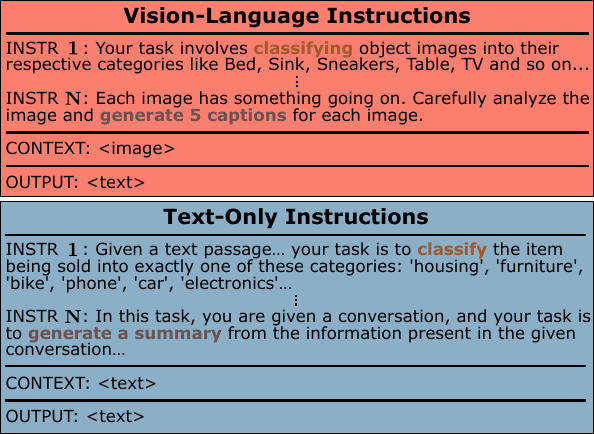}
        \vspace{0.01mm}
        \caption{
            Examples of instructions across modalities that share similar goals.
        }
        \label{fig:results}
    \end{subfigure}
    \caption{Similarity between \languageonly\ and \visionlanguage\ instruction tuning data shown both (a) quantitatively with similarity scores and (b) qualitatively with examples. 100k instructions are sampled from the \supernatural~\citep{wang2022super} and \visionflan~\citep{Xu2024VisionFlanSH} datasets and embedded by a pretrained sentenceTransformer, all-mpnet-base-v2~\citep{mpnet}. The red vertical line denotes the mean score. We then randomly sample and display two instructions with high cosine similarities (0.53 \& 0.38). }
    \label{fig:motivation}
\end{figure*}

\section{Approach}
\method\ views vision instruction following abilities as a natural extension of \languageonly\ abilities, a transfer that can occur due to the extensive multimodal pretraining used in MLLMs.

We begin by motivating our method through an empirical analysis of similarities in \languageonly\ and \visionlanguage\ instruction tuning data, which leads to our hypothesis that \languageonly\ data can largely replace \visionlanguage\ data to improve performance on general tasks.
Then, we detail our training, following the standard design of existing instruction tuning methods~\citep{flan, multiinsturct, instructblip, llava} in four stages: selecting training data, formatting the data with instructions, fine-tuning a pretrained MLLM on the training set (Sec.~\ref{training-details}), and evaluating the instruction tuned model on standard academic benchmarks in the zero-shot setting (Sec.~\ref{evaluation-tasks}).

\subsection{Natural Correspondence between \languageonlycap\ and \visionlanguagecap\ Instructions}
\label{similarity-text-vision}
While the image-text and the text-only distribution of instructions significantly differ from each other, we observe shared semantics and structure on the task level when comparing wild instruction-response pairs in both modalities.

\paragraph{Semantic Similarity} We study two comprehensive large-scale instruction tuning datasets with one from each modality, namely \supernatural~\citep{wang2022super} and \visionflan~\citep{Xu2024VisionFlanSH}, which are representative of common structures and tasks. We show \visionlanguage\ and \languageonly\ tasks are similar by randomly sampling 100k instances from each dataset and examining the distribution of the cosine similarities between embedded instructions as shown in Figure~\ref{fig:motivation}(a). A significantly non-negative mean cosine distance provides evidence that the tasks performed in either domain are somewhat similar, based on the belief that tasks are defined by the instructions. 
Additionally, there is a small yet nonzero chance to even see a pair of tasks that are comparable with high similarities ($>$0.3) in the language and vision domain. To qualitatively demonstrate this, in Figure~\ref{fig:motivation}(b) we show two pairs of semantically similar instructions from each datasets with a similarity score of 0.53 and 0.38, respectively. While the first example is a classical classification task, the second requests a concise representation of the context, where the context may be a text paragraph or an image. We reason that if the ability to describe a casual conversation is acquired, the ability to caption an image can be readily obtained.

\paragraph{Structural Similarity} The well-established problem of solving zero-shot tasks can be split into a user prompt followed by a model's response for both modalities. While some text-only tasks appeal to a model's internal knowledge, such as ARC~\citep{arc}, the task of open-book question answering is analogous to vision question answering in the sense that additional inputs are provided to serve as the reference where the final answer is derived. If the vision and the text modalities are well aligned, it makes sense for a model to easily refer to the details in an image as the image tokens are no different than the native word tokens in its embedding space.

\label{sec:training}
\subsection{Training Details}
Our approach, \method, is simple, changing the dataset composition across modalities compared to traditional MLLM instruction tuning. We finetune a multimodal pretrained LLM in the FLAN-style~\citep{flan} and further train on a small portion of vision instruction data (compared to the number of text-only instances) to adapt the model to \visionlanguage\ 
 queries. While mainstream methods, including LLaVA~\citep{llava} and Cambrian-1~\citep{cambrian-1}, also include some text-only examples in their vision instruction tuning dataset, their primary goal has been providing language as a form of regularization to prevent catastrophic forgetting. Our method differs by approaching vision instruction tuning from the other way around: we build strong language-only instruction-following abilities to build a robust knowledge base, and then introduce a small number of vision instances solely for grounding and domain adaptation. To demonstrate that adjusting the data composition alone is a viable substitute for vision-heavy instruction tuning, we use a fixed size budget and shared data sources for all our experiments, thus controlling the effect of longer training sessions and variable data quality.

\label{training-details}
\paragraph{Dataset Selection}
\label{dataset-selection}
Inspired by the similarity in instruction tuning across modalities, we use the same two diverse datasets to train with, encompassing a multitude of tasks in each modality. 
For \languageonly\ data we sample from the over 1600 tasks in \supernatural~\citep{wang2022super}, while for \visionlanguage\ data we sample from the 187 tasks in \visionflan~\citep{Xu2024VisionFlanSH}.
This gives us ample coverage across many \languageonly\ and \visionlanguage\ tasks.
For all of our experiments, we use a fixed data budget of \totaltraininstances\ instances, which can come from either \supernatural\ or \visionflan\ depending on the setting.

\paragraph{Models and Multimodal Pretraining} 
We follow the architecture design of LLaVA \citep{llava1.5} that connects a visual encoder with a projector that enables the LLM to use the outputs of the visual encoder to process image inputs in addition to texts.
We choose CLIP-ViT-L/14@336 \citep{clip} and a two-layer MLP with GELU activation as the visual encoder and the projector, respectively. We select the base LLMs as \llama{3}~\citep{meta2024llama32} and \llama{8}~\citep{grattafiori2024llama3}, both the non instruct versions.
We conduct multimodal pretraining for both models on LLaVA-Pretrain-558K using the same hyper-parameters as in~\citet{llava1.5}. 
These models are then finetuned on our language-heavy training dataset for one epoch using a global batch size of 128, a cosine learning schedule, a learning rate of 2e-5, a warm-up ratio of 0.03, and no weight decay.
Both the visual encoder and LLM are frozen throughout the pretraining session while the parameters in the MLP projector are updated. 
After pretraining, the visual encoder and the projector function as a visual tokenizer that turns an image into tokens compatible with the LLM.

\paragraph{Instruction Tuning}
To test our instruction tuning methodology, we finetune MLLM checkpoints using a controlled mixture of \languageonly\ and \visionlanguage\ data, focusing on the former.
This is because language, rather than vision, remains the primary medium for users to interact with models when they specify their needs. 
In contrast, most existing multimodal instruction tuning approaches prioritize vision-language data and include language-only tasks merely to mitigate forgetting. ~\citep{llava1.5, qwen-vl, ye2023mplug, luo2024cheap, cambrian-1}.
These approaches require many more training tokens and rely on a greater number of \visionlanguage\ datasets.
See Table~\ref{app:language-training-composition} in Appendix~\ref{related-work-composition} for the percentage of \languageonly\ data included during instruction tuning for various state of the art MLLMs.
Current instruction tuning mixtures across models vary substantially in language content, yet few of these design choices are grounded in systematic empirical comparison.
Our method systematically tests the effectiveness of the composition of instruction tuning data by modality, then anchors in a shift in perspective, treating \textbf{language as the foundation} in instruction tuning.



\subsection{Evaluation Tasks}
\label{evaluation-tasks}
Our evaluation suite covers diverse \languageonly\ and \visionlanguage\ tasks for zero-shot evaluation that are not seen during training. 
The \languageonly\ benchmarks include \textbf{Commonsense understanding}, \textbf{Reasoning}, \textbf{Reading comprehension} and \textbf{Scientific knowledge testing}. Similarly, the selected \visionlanguage\ benchmarks primarily test \textbf{Scene Understanding} and \textbf{Image Reasoning}. 
Notably, MMLU~\citep{hendrycks2020measuring}, MMMU~\citep{MMMU}, and MME~\citep{MME} are large multidisciplinary benchmarks covering wide domains. 
We craft suitable instruction templates for each dataset in the same way as for the training datasets, using the same collection of instruction prompts.
The final evaluation collection includes \numtextevaltasks\ \languageonly\ datasets and \numvisionevaltasks\ \visionlanguage\ datasets. 
The answer types cover short-response, multiple-choice, and true/false questions. 
Appendix~\ref{dataset-summary} provides a brief description of each dataset.

\begin{table*}[!ht]
\centering
\resizebox{\linewidth}{!}{%
\begin{tabular}{ll|cccccc}
\hline
\multicolumn{2}{c|}{}                       &  \multicolumn{6}{c}{Vision Benchmarks} \\ 
\textbf{Models}         & \textbf{Method}   & POPE	             & ScienceQA-IMG	    & MMMU           & MME	               & MMBench	      & Avg. \\ 
\hline
                        & Pretrain          & 66.67$^*$	         & 43.73	            & 26.44	         & 700$^*$	           & 51.10	          & 42.59 \\
\llama{3}               & \llavamix\             & 80.10	             & 64.65	            & 29.00	         & 1293.56	           & \textbf{67.71}	  & 57.53 \\
                        & \cambrianmix\        & 81.90	             & \textbf{65.94}	    & 28.67	         & 1367.38	           & 67.48	          & 58.57 \\
                        & \method           & \textbf{83.17}	 & \textbf{65.94}	    & \textbf{29.33} & \textbf{1405.53}	   & 67.01	          & \textbf{59.13} \\
\hline 
                        & Pretrain          & 66.67$^*$		     & 63.81	            & 27.67	         & 700$^*$	           & 62.81	          & 49.61\\
\llama{8}               & \llavamix\             & 79.90	             & 67.97	            & 30.89	         & 1354.52	           & 70.29	          & 59.49\\
                        & \cambrianmix\        & \textbf{82.57}	 & 70.55	            & \textbf{36.00} & 1408.02	           & \textbf{73.50}	  & \textbf{62.58} \\
                        & \method           & 81.84	             & \textbf{71.15}	    & 34.44	         & \textbf{1436.83}	   & 72.51	          & 62.25\\
\hline
\end{tabular}
}
\caption{
Zero-shot results on the held-out \visionlanguage\ datasets for \llama{3} and \llama{8}. 
We compare Pretrain, \llavamix, \cambrianmix, and \method\ (ours). $^*$ denotes that the pre-trained models fail to generate meaningful responses other than all "yes" or "no". ScienceQA~\citep{sciq} is included in \visionflan\ but excluded in experiments. The MME scores are normalized by dividing by the maximum value (2800) when computing the average.
}
\label{tab:heldout-vision-performance}
\end{table*}

\begin{table*}[!ht]
\centering
\resizebox{\linewidth}{!}{%
\begin{tabular}{ll|ccccccccc}
\hline
\multicolumn{2}{c|}{}                   &  \multicolumn{9}{c}{Language Benchmarks} \\ 
\textbf{Models}     & \textbf{Method}   & ARC-E             & ARC-C             & CommensenseQA         & PIQA            & RACE            & BoolQ         & CosmosQA       & MMLU             & Avg. \\ 
\hline
                    & Pretrain          & 62.42	            & 42.41	            & 63.72	                & 76.77	          & \textbf{70.37}	& 62.91	        & \textbf{67.77} & 24.09	        & 58.81  \\
\llama{3}           & \llavamix\             & 69.40	            & 43.34	            & 58.39	                & 78.40	          & 58.57	        & 68.93	        & 47.57	         & 44.65	        & 58.66  \\
                    & \cambrianmix\        & \textbf{71.68}    & 46.25	            & 60.85	                & \textbf{79.27}  & 67.98	        & \textbf{71.59}& 59.40	         & 48.39	        & 63.18  \\
                    & \method\          & 71.30	            & \textbf{46.93}	& \textbf{66.18}	    & 79.11	          & 70.27	        & 68.44	        & 64.76	         & \textbf{49.03}	& \textbf{64.50}  \\
\hline 
                    & Pretrain          & 71.09	            & 50.00	            & 70.19	               & 80.14	          & 79.41	        & 64.89	        & 76.65	         & 39.79	        & 66.52\\
\llama{8}           & \llavamix\             & 72.60	            & 48.81	            & 66.20	               & 79.43	          & 71.44	        & 75.38	        & 59.53	         & 50.51	        & 65.49  \\
                    & \cambrianmix\        & 72.80	            & 48.81	            & 68.88	               & 80.03	          & 74.22	        & 77.22	        & 64.42	         & 55.69	        & 67.76  \\
                    & \method\          & \textbf{74.79}	& \textbf{50.17}	& \textbf{73.05}       & \textbf{81.23}	  & \textbf{79.91}	& \textbf{78.53}& \textbf{76.68} & \textbf{58.18}	& \textbf{71.57}  \\
\hline
\end{tabular}
}
\caption{
Zero-shot results on the held-out \languageonly\ datasets for \llama{3} and \llama{8}. 
We compare Pretrain, \llavamix, \cambrianmix, and \method\ (ours).
}
\label{tab:heldout-language-performance}
\end{table*}


\section{Experiments}


In this section, we show that \method\ is both more effective and training efficient compared to the pretrained MLLMs as well as state of the art multimodal instruction tuning mixtures across all the tasks we evaluate. 
Additional details of the training and experimental setup are described in Appendix~\ref{additional-training-details}.

\subsection{Main Results}
We compare various instruction tuning methods built upon our multimodal pretrained \llama{3} and \llama{8}.
We include the following settings, all using our specified training methodology, only varying composition:
(1) Pretrain: The MLLM after multimodal pretraining with no instruction tuning.
(2) \llavamix\ and \cambrianmix: We use our training dataset along with the multimodal instruction tuning mixture recipes of LLaVA~\citep{llava1.5} and Cambrian-1~\citep{cambrian-1}, i.e., with 6\% and 25\% \languageonly\ instruction data, respectively.
(3) \method: Our text-first instruction tuning method with a composition heavily favoring (75\%) \languageonly\ data.

\paragraph{Cross-Task Generalization} 
We report the scores of pretrained MLLM and instruction tuned models on \numevaltasks\ benchmarks in Tables~\ref{tab:heldout-vision-performance} and \ref{tab:heldout-language-performance}, respectively. Compared with \cambrianmix, \method\ yields the best performance in the 3B setting and matches the best score in the 8B setting, only falling behind by 0.33\%, despite being trained on less than half of the images. The competitive vision performance shows effective cross-modal transfer. MLAN consistently improves performance on knowledge-intensive tasks such as MMLU, CosmosQA, and ARC-C, demonstrating stronger internal knowledge retention compared to vision-heavy baselines.

\paragraph{Knowledge Erosion}
We note that both \llavamix\ and \cambrianmix\ suffer from catastrophic forgetting, especially on CommonsenseQA~\citep{commonsenseqa} and CosmosQA~\citep{cosmosqa}, showing performance degradations up to 5.3\% and 20.2\%. However, \method\ is more resilient against forgetting. In the only case where its performance decreases in CosmosQA, the decline is significantly smaller than other models (3.1\% vs. 20.2\% \& 8.37\%). On all other benchmarks, including vision, our method shows a solid positive gain. Such an observation unveils an asymmetrical interaction between vision and text modalities, where the text ability is more susceptible to forgetting, but the vision ability generally benefits from language-based tuning. This trend is explored again in Section~\ref{sec:scaling}.

\begin{table}[ht]
    \centering
    \resizebox{\columnwidth}{!}{%
    \begin{tabular}{l r@{}l}
        \hline
        \textbf{Method} & \multicolumn{2}{c}{\textbf{Number of Tokens}} \\
        \hline
        \languageonlycap\ IT       & 37,906,142    & \\
        \method                    & 60,112,680    & \\
        \cambrianmix              & 101,480,339   & \textcolor{red}{$\uparrow$68.8\%} \\
        \llavamix                 & 117,220,955   & \textcolor{red}{$\uparrow$95.0\%} \\
        Full \visionlanguagecap\ IT & 122,054,758   & \textcolor{red}{$\uparrow$103.0\%} \\
        \hline
    \end{tabular}%
    }
    \caption{All token counts for various training settings with \totaltraininstances\ total instances. The percentage score indicates the size increase relative to the \method\ setting.}
    \label{tab:lit-vit-token-counts}
\end{table}

\subsection{Training Efficiency}
A major advantage of our method is that it significantly reduces the computational cost measured by the number of training tokens processed by the base LLM compared to vision-based instruction tuning. Table \ref{tab:lit-vit-token-counts} details the number of training tokens, including those in the visual prefix. Visual inputs drastically increase the training burden as an image is converted to hundreds of visual tokens (576 tokens with CLIP-ViT-Large-patch14@336~\cite{clip}) before being processed along with regular text tokens. 
Therefore, \method\ stands out as a more efficient vision instruction-tuning approach that avoids excessive instruction tuning on images.

\label{sec:scaling}
\subsection{Knowledge Transfer Curve}
\begin{figure}[htbp]
    \centering
    \includegraphics[width=\columnwidth]{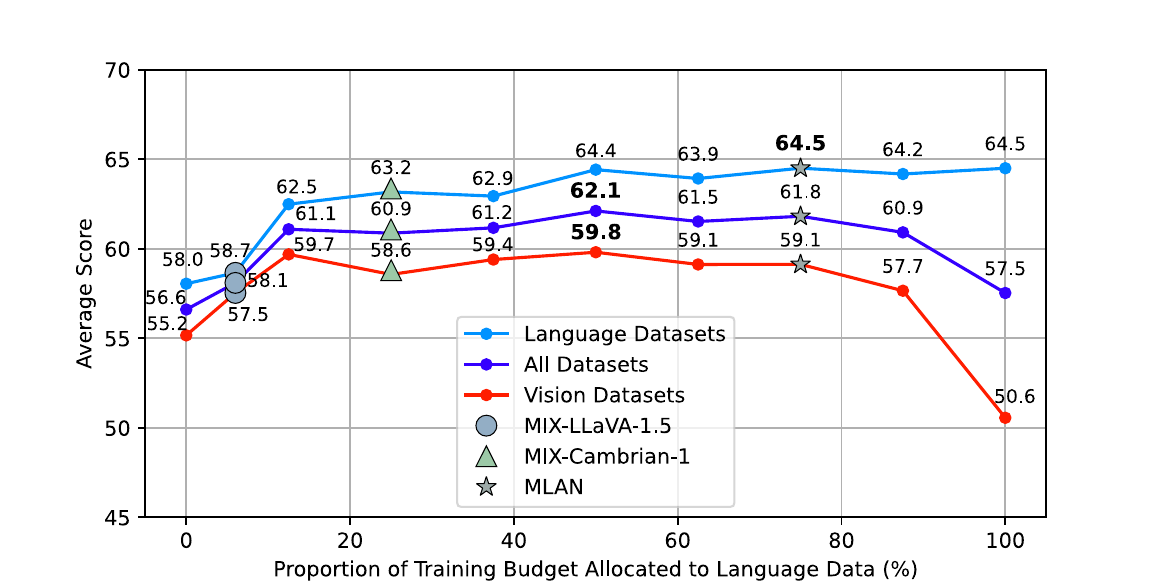}
    \caption{
    Average scores on \llama{3} based MLLMs with respect to the percentage of language data mixed in. The percentage denotes the amount of language data.
    }
    \label{fig:lineplot}
\end{figure}

To better understand the role of language, we perform a controlled study by varying the proportion of language-only data in the instruction tuning mixture, increasing it in 12.5\% increments. We show the performance of \llama{3}-based MLLMs with different amounts of language instruction data in Figure~\ref{fig:lineplot}. 
Notably, we observe that even a small amount of language data (12.5\%) leads to a sharp increase in both text and vision performance, suggesting that foundational knowledge acquired through language tuning quickly transfers across modalities.
As the proportion of language data increases further, text performance continues to improve, whereas vision performance peaks and then slightly declines.
Full vision-language tuning fails to match the peak vision performance achieved with a balanced mix, indicating that language-based knowledge is not only transferable but also essential for efficient vision instruction tuning. 
This analysis reinforces our central claim: language acts as a scaffold for multimodal reasoning, and a moderate inclusion of vision data is sufficient for grounding.

\textbf{\begin{table}[ht]
    \centering
    \scriptsize
    \begin{tabular}{l l r r}
        \hline
        \textbf{Base LLM} & \textbf{Variant} & \textbf{Text Avg.} & \textbf{Vision Avg.} \\
        \hline
        \llama{3}
            & +MLAN            & 64.50 & 59.13 \\
        \rowcolor{green!15}
            & +Instruct LLM    & \textbf{67.74} & \textbf{60.98} \\
        \rowcolor{yellow!15}
            & -25\% tasks      & 65.68 & 55.71 \\
        \rowcolor{yellow!15}
            & -50\% tasks      & 65.98 & 55.73 \\
        \rowcolor{yellow!15}
            & -75\% tasks      & 66.35 & 56.79 \\
        \hline
    \end{tabular}
    \caption{Ablation study on Llama-3.2-3B with different instruction tuning variants and fewer tasks.}
    \label{tab:llama32-ablation}
\end{table}
}


\subsection{Additional Instruction Tuning Factors}

\paragraph{Instruction Tuned Base Models}
\label{existing-instr-models}
We use base (non instruction tuned) models in our experiments to show the impact of \languageonly\ data while controlling the amount of text instruction tuning. 
However, mainstream vision instruction tuning methods mostly choose instruction-tuned (chat) models as the default LLM backbone~\citep{llava,nvlm,vila}.
Table~\ref{tab:llama32-ablation} shows that finetuning the instruction tuned variant instead of the pretrained model readily boosts both text and vision performance by 2-3\%, even when we continue to emphasize \languageonly\ data in the visual instruction tuning phase.
\nick{Let's clarify that the instruct LLM is still with MLAN mixture?}
This provides more evidence that the text-first approach throughout training is beneficial. A possible explanation for this is that the model adapts to the instruction following format and eliminates the distributional shift from the pretraining to the instruction tuning corpus.

\paragraph{Task Diversity within Datasets}
\label{dataset-diversity}
Prior work has emphasized the importance of diversity within instruction tuning datasets~\citep{liEagle2Building2025, Xu2024VisionFlanSH, flan}. We conduct a controlled finetuning experiment by reducing the proportion of included tasks (25\%, 50\%, 75\%, 100\%) while keeping the total number of training instances fixed.
Surprisingly, text performance slightly improves with fewer tasks, peaking at 25\%, while vision performance only improves with full task coverage.
This suggests that task diversity does not uniformly benefit all modalities: some tasks may be less helpful, and that over-diversification may dilute useful supervision, especially for language.
\nick{Did you end up doing any more of the other task \% experiments? In line with the ACL comments may be nice to have values for 25 or 50\%}

\begin{table}[ht]
    \centering
    \scriptsize
    \resizebox{\columnwidth}{!}{%
    \begin{tabular}{llllll}
        \hline
        \textbf{Base LLM} & \textbf{PT} & \textbf{IT} & \textbf{Text Avg.} & \textbf{Vision Avg.} \\
        \hline
        \multirow{4}{*}{Llama-3.2-3B} 
            & LLaVA & \visionflan\       & 55.14 & 57.61 \\
            &       & Super-Natural      & \textbf{64.89} & 46.95 \\
            & ShareGPT4V & \visionflan\  & 58.05 & \textbf{58.26} \\
            &           & Super-Natural  & 64.48 & 50.20 \\
        \hline
        \multirow{4}{*}{Llama-3.1-8B} 
            & LLaVA & Vision Flan        & 63.08 & 54.77 \\
            &       & Super-Natural      & \textbf{72.05} & 52.55 \\
            & ShareGPT4V & Vision Flan   & 58.27 & \textbf{56.84} \\
            &           & Super-Natural  & 71.76 & 50.76 \\
        \hline
    \end{tabular}%
    }
    \caption{Average performance across different vision pretraining (PT) and instruction tuning (IT) strategies.}
    \label{tab:pretraining-ablation}
\end{table}

\subsection{Interaction between Pretraining and Single-Modal Instruction Tuning}
Before visual instruction tuning, the vision pretraining step aims to align the text and vision modalities. 
Increasing pretraining data has been shown to increase post instruction tuning performance given the same corpus~\citep{mm1}, but changes in pretraining data have been shown to have minimal effects~\citep{cocchiLLaVAMOREComparativeStudy2025}.
To investigate how the pretraining dataset affects instruction tuning on various modalities, we conduct experiments using single-modality instruction tuning datasets on another pretraining dataset (Table~\ref{tab:pretraining-ablation}).
Although we expect models to benefit from higher quality samples and longer training sessions due to ShareGPT4V~\citep{sharegpt4v}, the results demonstrate that this is only consistently true when the model is finetuned with vision-text instruction data. More vision pretraining has a mixed effect on the text performance, boosting the 3B model's text score while hurting the 8B model's performance. Additionally, scaling up the model size effectively increases the text scores but leaves the vision scores roughly on the same level.
\nick{Clarify more the differences between LLaVA and ShareGPT4V so it is clear the latter is expected to be better}

\paragraph{Diversity in Training Data}

\begin{figure}[htbp]
    \centering
    \includegraphics[width=\linewidth]{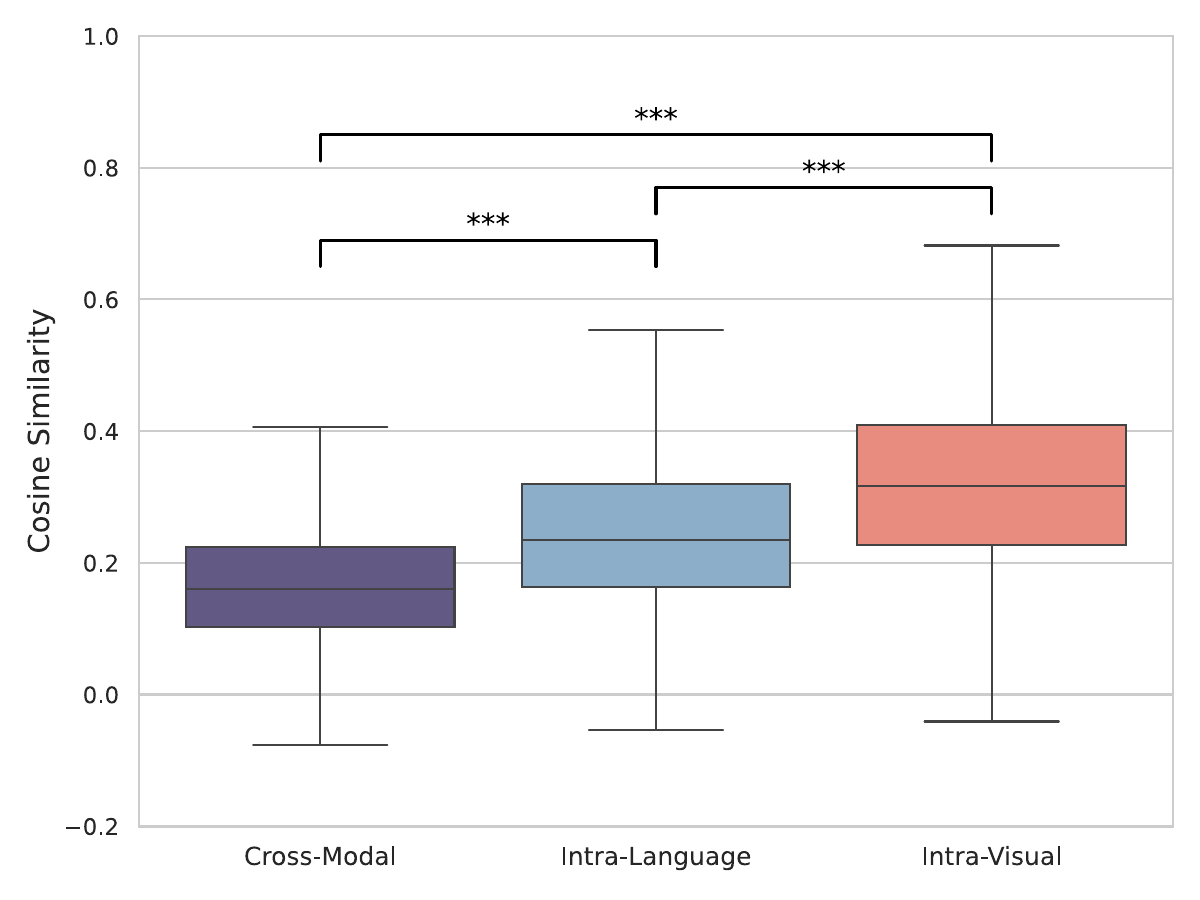}
    \caption{
        Distribution of the cosine similarity of random question pairs sampled in the language and \visionlanguage\ settings. The stars (***) indicate significant differences $(p<0.001)$ between the mean similarity supported by the t-test.
    }
    \label{fig:diversity2}
\end{figure}

In Section~\ref{similarity-text-vision}, we explored the similarity between instruction tuning using \languageonly\ and \visionlanguage\ data. 
We now compare the mean cosine distances in two intra-dataset and one inter-datasets settings.
Figure \ref{fig:diversity2} reports the mean cosine similarities. 
The \visionlanguage\ appears more homogeneous, with a higher mean, while the language data is more diverse. 
This observation aligns with the fact that vision-language datasets typically contain fewer distinct task types and tend to emphasize perceptual grounding, whereas language-only corpora encompass a broader spectrum.
Importantly, the similarity scores between language-only and vision-language instructions are comparable to those within the language-only set, suggesting that diverse linguistic tasks inherently support better generalization—even across modalities.
This could imply that language data, at least in our training data, better generalizes to vision datasets thanks to greater heterogeneity.
Notably, though we use a diverse set of \languageonly\ and \visionlanguage\ data, there is still a gap between the similarities, meaning \languageonly\ data that aligns better with \visionlanguage\ can likely be constructed, which may improve performance even more.

\section{Related Work}
Multimodal large language models (MLLM) are language models endowed with the ability to use multiple modalities, such as images, videos, and audio~\citep{gpt4o, meta2024llama32, team2023gemini, openai2024gpt4technicalreport,rubenstein2023audiopalmlargelanguagemodel, zhang-etal-2023-speechgpt, ataallah2024minigpt4videoadvancingmultimodalllms,baiQwen25VLTechnicalReport2025,liLLaVAOneVisionEasyVisual2024,liuNVILAEfficientFrontier2025, agrawalPixtral12B2024, deitkeMolmoPixMoOpen2024,chenExpandingPerformanceBoundaries2025}.
The most widely adopted are vision enhanced LLMs, where many design choices are already extensively studied~\citep{llava1.5, mm1, vila, idefics2, cambrian-1, prismaticvlm, cocchiLLaVAMOREComparativeStudy2025, liEagle2Building2025}.
A prevalent approach to building such MLLMs links pretrained visual encoders~\citep{clip, dinov2} to LLMs~\citep{llama2, vicuna, vicuna2023} via an adapter, thus transforming deep image features into soft prompts for the base LLM. 
In our work, we focus on one of the simplest yet high-performing and widely adopted MLLMs, using only a multi-layer perceptron as the adapter~\citep{llava, llava1.5, liu2024llavanext, li2024llavanext-ablations, palme, vila, lynx}.
Inspired by the success of instruction tuning in LLMs in zero-shot generalization~\citep{flan,wang2022self,zhang2023instruction, ouyang2022training}, following a pretraining step for \visionlanguage\ feature alignment, there is a multimodal instruction tuning step to improve zero-shot performance on multimodal tasks~\citep{multiinsturct, li2024vision}.
Notably, InstructBLIP~\citep{instructblip} and LLaVA~\citep{llava} transform existing datasets into multimodal instructions using manual templates and synthetic data, a practice expanded upon in subsequent work~\citep{cambrian-1, politeflamingo, vila}.
Further work investigates how instruction tuning varies under different settings, e.g., how different components of the MLLM should learn differently during instruction tuning~\citep{wu2024commit} and how instruction tuning works in a continual learning setting with many new tasks~\citep{chen2024coin}.
However, there lacks a comprehensive set of experiments that varies the composition of each modality in instruction tuning.

Though the primary goal of multimodal instruction tuning is to improve \visionlanguage\ performance, \languageonly\ data is often included in both pretraining~\citep{mm1, vila} and finetuning~\citep{llava1.5, kosmos-1, qwen-vl, ye2023mplug, ye2024mplug, luo2024cheap, vila, cambrian-1, nvlm, baiQwen25VLTechnicalReport2025, liEagle2Building2025, zhangMM15MethodsAnalysis2024, zhang2024wings} to prevent catastrophic forgetting and improve language performance.
Many such papers disregard the impact of finetuning with \languageonly\ data on vision performance, focusing solely on language performance when ablating \languageonly\ data away, though there are notable exceptions~\citep{kosmos-1, ye2023mplug, ye2024mplug, vila, nvlm, zhangMM15MethodsAnalysis2024}. 
In these cases, there is modest evidence of transferability between modalities, where finetuning on both language and vision data exhibits about equal or better performance than training on one modality alone.
However, in each of the existing work that finetune with \languageonly\ data alongside vision data, this performance boost is achieved by increasing the dataset size without consideration of how such data will increase the training cost (with the exception of~\citet{zhangMM15MethodsAnalysis2024}, which only tests with a low amount of \languageonly\ data).
Hence, even though better performance is obtained when increasing the dataset size to train on \languageonly\ data, the instruction tuning step is more costly.

Due to the general cost of instruction tuning a MLLM, many approaches aims to decrease the cost of instruction tuning in the multimodal setting. These primarily include using lightweight adapters to decrease the number of parameters~\citep{luo2024cheap, llama-adapter,liuReImaginingMultimodalInstruction2025} and choosing a subset of the training data using the MLLM itself or other methods~\citep{chen2024your, wei2023instructiongpt, lee2024concept, liu2024less, safaeiFilterImagesFirst2025,biPRISMSelfPruningIntrinsic2025}.
A simpler way to decrease the cost is to instruction tune with a focus on \languageonly\ data. 
Since training on language instruction data is cheaper than training on the same number of vision instances, and language is foundational to the functioning of MLLMs, we focus on such a language-based approach.
\section{Conclusion}
We present \method, a language-based multimodal instruction tuning strategy for MLLMs that enhances zero-shot generalization and promotes effective knowledge transfer across modalities.
We demonstrate—through controlled ablations under fixed training budgets—that language-based tuning establishes a robust knowledge foundation, even for tasks requiring visual understanding.
Crucially, \method\ achieves strong performance on both language and vision benchmarks while significantly reducing reliance on image supervision.
Our results show that language is not only sufficient but essential for efficient and generalizable multimodal learning.
With \method, we hope to bring attention to the importance of language in MLLMs in visual instruction tuning, which we believe can be used in future work to improve training efficiency and performance.

\newpage
\section{Limitations}
Our experiments are performed on models with the same multimodal architecture and pretraining procedure, not accounting for more advanced architecture or large-scale multimodal pretraining. 
Though we evaluate on a comprehensive set of \visionlanguage\ benchmarks, we do not evaluate on specialized out of distribution tasks like OCR or captioning, focusing only on general tasks where the transferability is motivated.
We invite future work to explore other methodologies to find where such specialized \languageonly\ and \visionlanguage\ tasks align.
Our analysis could also use experiments testing how instruction tuning varies when different tasks are trained on versus held-out, or on sequential finetuning versus sampling \languageonly\ and \visionlanguage\ data.
Furthermore, the instruction tuning experiments have the same data budget of \totaltraininstances\ instances, while existing instruction tuning data may contain hundreds of thousands or even multi-million instances, which we leave to future work.
\bibliography{custom}

\appendix

\section{Additional Implementation Details}
For language-based instruction tuning, we use our carefully crafted dataset with tasks across modalities. To avoid data contamination, only the train split of each dataset is used for finetuning, and the test split, or the validation split if the test split is not publicly available, is reserved for evaluation.
Similar to various multimodal instruction tuning work~\citep{multiinsturct, instructblip}, we select unseen datasets of both modalities for evaluation. They are used to quantify performance in a general setting. 

We maintain a fixed data budget of \totaltraininstances\ instances throughout the training sessions. All training instances are sampled from \supernatural\ and \visionflan, according to the designated ratio. For the former, to prevent overfitting to a specific task, we sample an equal number of instances from every task. For the latter, since ScienceQA~\citep{sciq} is included in the training set, we manually remove them for evaluation purposes so there is no contamination.
For finetuning, we apply the same chat template to all models in the following format: "USER:$<$query$>$ASSISTANT:$<$response$>$". The same prompt is used to format inputs during evaluation. 

\section{Additional Training Details}
\label{additional-training-details}
We finetune pretrained MLLMs on the \languageonly\ data and denote those with a 75\% \languageonly/25\% \visionlanguage\ split as \method.
Acknowledging the recent trend of including a small portion of \languageonly\ data into vision instruction tuning data, we establish two additional baselines by finetuning on two separate versions of our training dataset that contain only 6\% and 25\% language instruction data, similar to the ratio in \citet{llava1.5} and \citet{cambrian-1}. For a fair comparison, we limit the total number of training sequences in all settings to \totaltraininstances\ samples from our training data. 

\section{Dataset Summary}
\label{dataset-summary}
In Tables~\ref{tbl:updated_dataset_summary} and \ref{tbl:dataset-description}  we provide information about all \numevaltasks\ benchmarks used for evaluation. Note that in the main body we present results on 13 datasets, as we do not combine ARC-E and ARC-C.

\begin{table*}[htbp]
\centering
\resizebox{\linewidth}{!}{%
\begin{tabular}{l|l|l|l|l|l}
\hline
\textbf{Dataset} & \textbf{Modality} & \textbf{Split} & \textbf{Answer Type} & \textbf{Dataset Type} & \textbf{Size} \\
\hline
ARC-Easy \citep{arc}          & Text    & Test        & Multiple Choice  & Held-out  & 2.2k  \\
ARC-Challenge \citep{arc}    & Text    & Test        & Multiple Choice  & Held-out  & 1.2k  \\
BoolQ     \citep{boolq}        & Text    & Validation  & True/False       & Held-out  & 3.2k  \\
CommonsenseQA  \citep{commonsenseqa}   & Text    & Validation        & Multiple Choice  & Held-out  & 9.7k  \\
PIQA     \citep{Bisk2020}           & Text    & Validation        & Multiple Choice  & Held-out  & 16.1k \\
MMLU      \citep{hendrycks2020measuring}        & Text    & Test        & Multiple Choice  & Held-out  & 14.0k \\
RACE       \citep{race}       & Text    & Test        & Multiple Choice  & Held-out  & 1.05k \\
CosmosQA     \citep{cosmosqa}     & Text    & Validation  & Multiple Choice  & Held-out  & 3.0k  \\
POPE        \citep{pope}      & Vision  & Test        & True/False       & Held-out  & 9.0k  \\
ScienceQA-IMG  \citep{sciq}    & Vision  & Test        & Multiple Choice  & Held-out  & 5.0k  \\
MMMU        \citep{MMMU}      & Vision  & Validation        & Multiple Choice  & Held-out  & 1.5k  \\
MME         \citep{MME}      & Vision  & Test        & True/False       & Held-out  & 2.8k  \\
MMBench      \citep{liu2024mmbench}     & Vision  & Dev        & Multiple Choice  & Held-out  & 5.2k  \\
\hline
\end{tabular}%
}
\caption{Overview of evaluation datasets.
\label{tbl:updated_dataset_summary}
}
\end{table*}

\begin{table*}[htbp]
\centering
\resizebox{\linewidth}{!}{%
\begin{tabular}{p{6cm}|p{14cm}}
\toprule
\textbf{Dataset}                     & \textbf{Descriptions}           \\ 
\midrule
CosmosQA \citep{cosmosqa}             & Questions require reasoning based on people's everyday narratives to deduce the causes and effects of pertinent events. \\ \hline
CommonsenseQA \citep{commonsenseqa}   & CommonsenseQA contains questions without context about understanding and relations between common objects.  \\ \hline
ARC \citep{arc}                       & ARC consists of grade-school level multiple-choice questions about understanding scientific concepts. Both easy and challenge splits are used. \\ \hline
RACE \citep{race}                     & Race contains questions about long paragraphs collected from K12  English examinations in China.  \\ \hline
BoolQ \citep{boolq}                   & BoolQ asks whether a statement about a given long context is correct. \\ \hline
MMLU \citep{hendrycks2020measuring}   & A benchmark testing multi-task language understanding across 57 subjects, assessing model performance on expert-level multiple-choice questions. \\ \hline
PIQA \citep{Bisk2020}                  & PIQA evaluates physical commonsense reasoning by selecting the most plausible solution to everyday scenarios. \\ \hline
MME \citep{MME}                       & MME is a multimodal benchmark for assessing cognition and perception capabilities of MLLMs across multiple domains with yes and no questions. \\ \hline
MMMU \citep{MMMU}                      & A multi-disciplinary benchmark testing on expert-level knowledge with vision and question queries. Questions types contain short response and multiple choice. \\ \hline
MMBench \citep{liu2024mmbench}        & A comprehensive multimodal benchmark that evaluates scientific knowledge with multiple choice questions. \\ \hline
POPE \citep{pope}                     & POPE asks to determine whether an object is present in the scene. We use adversarial, popular, and random splits for evaluation.   \\ \hline
ScienceQA \citep{sciq}                & ScienceQA contains both \visionlanguage\ and \languageonly\ questions about scientific concepts. We use all questions to test the overall ability of our models.\\
\bottomrule
\end{tabular}
}
\caption{Short descriptions for the evaluation benchmarks in our study.}
\label{tbl:dataset-description}
\end{table*}

\section{Additional Related Work}
Our work focuses on choosing a simple multi-layer perception as the adapter in LLaVA \cite{llava, llava1.5}.
In contrast, BLIP-2 \cite{blip2} and Flamingo \cite{flamingo} design attention-based modules to attentively pool visual features, among a variety of other choices that combine existing methods or create new ones~\citep{zhu2023minigpt, internVL, idefics2}.
To train the model, most often there is a pretraining step focusing on aligning the multimodal features with a modality connector~\citep{yin2023survey}, though some models are trained from scratch~\citep{kosmos-1, florence-2}. 
A main design choice in MLLMs is whether to freeze or unfreeze the LLM during finetuning. 
Unfreezing the LLM effectively prevents catastrophic forgetting by maintaining \languageonly\ performance~\citep{meta2024llama32, palme, flamingo}, but results in worse \visionlanguage\ performance~\citep{vila, nvlm}. In our work, we show that with an unfrozen LLM, training on a strong language-based dataset on a fixed data budget improves performance across modalities.
To evaluate MLLMs, there are a wide variety of \visionlanguage\ tasks~\citep{multiinsturct, instructblip, cambrian-1}. However, Cambrian-1~\citep{cambrian-1} demonstrate that certain \visionlanguage\ datasets, including some we used (AI2D and RealWorldQA), exhibit only a minor drop in performance of around 5\% if vision is disabled, suggesting that current \visionlanguage\ evaluations may be more language-focused. Though there is a need for more vision-centric analysis, this emphasizes how important language is in many vision tasks, a fact central to our work.

\subsection{\languageonlycap\ Data in Existing Work}
\label{related-work-composition}
Table~\ref{app:language-training-composition} lists dataset sizes as well as the splits between \visionlanguage\ and \languageonly\ data in popular models that use both.
We note that most models instruction tune with a majority of \visionlanguage\ data, with the exception of Kosmos-1~\citep{kosmos-1} being a model that uses language alone, though it has an extensive pretraining step that differs from the simple MLLM adapter paradigm.
Ultimately, many papers do not share their overall composition, and the ones that do vary greatly. We hope our work prompts the community to be more open in sharing their results and to do more work finding an effective and efficient ratio that can be used successfully across models.


\begin{table*}[ht]
    \centering
    \resizebox{\textwidth}{!}{
    \begin{tabular}{lccc}
        \toprule
        \textbf{Name} & \textbf{\languageonlycap\ Size} & \textbf{Total Size} & \textbf{\languageonly\ (\%)} \\
        \midrule
        LLaVA-1.5~\citep{llava1.5} & 40k & 665k & 6.0\% \\
        QwenVL~\citep{qwen-vl} & N/A & 350k & N/A \\
        QwenVL2.5~\citep{baiQwen25VLTechnicalReport2025} & $\sim$1M & $\sim$2M & 50\% \\
        NVLM~\citep{nvlm} & N/A & N/A & N/A \\
        VILA~\citep{vila} & 1M & N/A & N/A \\
        mPLUG-Owl~\citep{ye2023mplug} & 242k & 392k & 61.7\% \\
        mPLUG-Owl2~\citep{ye2024mplug} & 558k & 1.23M & 45.4\% \\
        PrismaticVLM~\citep{prismaticvlm} & 40k & 665k & 6.0\% \\
        MM1~\citep{mm1} & N/A & 1.45M & N/A \\
        MM1.5~\citep{zhangMM15MethodsAnalysis2024} & -- & -- & 10\% \\
        Kosmos-1~\citep{kosmos-1} & 122.5k & 122.5k & 100\% \\
        LaVIN~\citep{luo2024cheap} & 52k & 204k & 25.5\% \\
        Cambrian-1~\citep{cambrian-1} -- Cambrian-7M & 1.68M & $\sim$7M & 23.8\% \\
        Eagle 2~\citep{liEagle2Building2025} -- Stage 1.5 & 4.75M & 21.6M & 22.0\% \\
        LLaVA-OneVision~\citep{liLLaVAOneVisionEasyVisual2024} -- Single-Image Data & 457.6k & 3.2M & 14.3\% \\
        \bottomrule
    \end{tabular}
    }
    \caption{Language instruction tuning dataset sizes in existing MLLMs. N/A means the number is either not presented in the paper or is unclear. A dash means the size is unclear.}
    \label{app:language-training-composition}
\end{table*}
\label{sec:appendix}

\end{document}